\documentclass[article,12pt]{elsarticle}

\usepackage[english]{babel}
\usepackage[T1]{fontenc}
\usepackage{lmodern}

\usepackage{amsmath,amssymb,bm}

\usepackage{multirow}
\usepackage{booktabs}
\usepackage{threeparttable}
\usepackage{threeparttablex}
\usepackage{colortbl}
\usepackage{pdflscape}
\usepackage{rotating}
\usepackage[ruled,vlined]{algorithm2e}

\usepackage{xcolor}
\usepackage{tcolorbox}

\usepackage{url}
\usepackage[hidelinks]{hyperref}

\usepackage{tikz}
\usetikzlibrary{shapes,arrows,positioning}

\usepackage{listings}
\usepackage[utf8]{inputenc}

\begin{document}

\begin{frontmatter}

\title{ARCADIA: Scalable Causal Discovery for Corporate Bankruptcy Analysis Using Agentic AI}

\author[1]{Fabrizio Maturo\corref{cor1}}
\ead{fabrizio.maturo@unimercatorum.it}

\author[2]{Donato Riccio}
\ead{donato@riccio.ai}

\author[1]{Andrea Mazzitelli}
\ead{a.mazzitelli@unimercatorum.it}

\author[2]{Giuseppe Bifulco}
\ead{giuseppe.bifulco@unimib.it}

\author[1]{Francesco Paolone}
\ead{francesco.paolone@unimercatorum.it}

\author[3]{Iulia Brezeanu}
\ead{iuliabrezeanu10@gmail.com}

\cortext[cor1]{Corresponding author}

\address[1]{Department of Economics, Statistics and Business, Universitas Mercatorum, Rome, Italy}
\address[2]{Department of Engineering and Science, Universitas Mercatorum, Rome, Italy}
\address[3]{Independent Researcher, Bucharest, Romania}

\begin{abstract}
Corporate bankruptcy prediction models typically identify correlational patterns but fail to uncover causal mechanisms driving firm failure. We introduce ARCADIA (Agentic Reasoning for CAusal DIscovery Algorithm), a framework that combines large language model reasoning with statistical validation to discover causal directed acyclic graphs (DAGs) for bankruptcy analysis. ARCADIA employs an iterative process where an LLM agent proposes theory-informed causal structures, evaluates them against statistical criteria, and refines the model based on diagnostic feedback. Evaluating against NOTEARS, GOLEM, and DirectLiNGAM baselines using Italian corporate data spanning several years prior to bankruptcy events, ARCADIA produces temporally coherent DAGs with no violations and minimal disconnected components, while baselines require substantial post-hoc pruning of invalid edges. ARCADIA maintains stable, interpretable graph sizes across varying problem complexities, whereas baselines exhibit high variance or collapse to trivial structures. ARCADIA prioritizes causal validity, essential for counterfactual analysis and regulatory compliance. This work demonstrates how agentic AI can bridge black-box prediction and explainable causal modeling, enabling better bankruptcy prevention and policy design in financial risk management.
\end{abstract}


\end{frontmatter}

\section{Introduction}

Research on corporate bankruptcy prediction has a long tradition in finance, dating back to ratio-based statistical models. Early empirical evidence showed that financial indicators carry substantial predictive power: Beaver \cite{beaver1966} demonstrated that selected accounting ratios could discriminate between failing and non-failing firms several years before default, while Altman \cite{altman1968} introduced the Z-score through multiple discriminant analysis, establishing one of the most influential models in the field. Subsequent developments moved toward probabilistic formulations, most notably Ohlson's O-score \cite{ohlson1980} and Zmijewski's probit-based distress model \cite{zmijewski1984}, which remain benchmarks in empirical corporate finance.
The increasing availability of high-dimensional data and advances in computation led to a shift from traditional statistical techniques to machine learning approaches. A large body of work shows that nonlinear classifiers and ensemble learning can substantially improve predictive accuracy in bankruptcy forecasting. Barboza et al.\ \cite{barboza2017} evaluate a wide range of machine learning models, including neural networks, SVMs and ensemble methods, reporting significant gains over classical specifications. Ensemble gradient-boosted trees have also proved effective in capturing complex interactions among financial variables \cite{zikeba2016}, while Support Vector Machines have been widely adopted, especially in the presence of imbalanced data and heterogeneous financial indicators \cite{sun2014}. More recent contributions leverage boosting-based architectures and additional information sets, such as text-based features, to improve robustness and generalizability \cite{chen2020, ha2023}.
At the same time, growing regulatory and managerial emphasis on explainability has motivated a parallel line of research on interpretable forecasting models. Rudin \cite{rudin2019} argues that transparent models are essential in high-stakes decision-making, where post-hoc explanations of black-box predictors may be unreliable. This view has led to the development of interpretable boosting models, feature-attribution techniques and hybrid approaches capable of combining predictive accuracy with clear economic interpretation. Liu et al. \cite{liu2022} provide an example by examining tree-based gradient boosting models through explainable machine learning tools tailored to financial distress prediction. In a similar spirit, recent work proposes fully explainable gradient boosting methods specifically designed for the Italian corporate context \cite{maturo2025anor}, showing that transparent machine learning models can perform competitively with black-box algorithms while providing clear insights into the drivers of corporate crises.

These conventional approaches – from discriminant analysis and logistic regression to modern black-box models – often rely on correlational patterns in financial ratios and market indicators. While such models can achieve high predictive accuracy, they offer limited insight into why a firm fails, and they struggle to generalize under changing economic conditions. In particular, correlational models cannot distinguish genuine causal drivers from spurious associations, making them brittle when used for policy analysis or stress testing. As Judea Pearl famously argued, "data alone" cannot establish causal relationships without assumptions about the data-generating process \cite{Pearl_00}. This limitation has tangible consequences in bankruptcy risk modeling: decision-makers lack a clear cause-and-effect narrative and cannot reliably answer counterfactual questions (e.g. \textit{Would a change in capital structure prevent default?}). Thus, there is a growing recognition that purely statistical bankruptcy prediction models, however sophisticated, leave a critical gap in explanatory power and robustness.

Causal inference offers a principled way to bridge this gap. By uncovering the directed acyclic graph (DAG) of causal relationships among variables (e.g. macroeconomic factors, firm financials, management decisions), one can move beyond prediction to understanding interventions and structural drivers of bankruptcy. However, implementing causal modeling in this domain is non-trivial. On one hand, fully data-driven causal discovery algorithms require strong assumptions (such as causal sufficiency and correct model specification) and large datasets to yield reliable graphs \cite{Glymour2019}. For example, the celebrated PC algorithm can, in theory, recover the skeleton of the causal graph from observational data, but it may falter if there are latent confounders or if sample sizes are limited \cite{Spirtes2000}. On the other hand, purely expert-driven causal modeling – where economists manually construct structural models – is labor-intensive and subjective, often feasible only for simplified scenarios. As a result, the application of causal DAGs in financial risk modeling remains limited. A recent survey observed that causal inference in banking and finance is still in its infancy, with relatively few studies to date \cite{Kumar2023}. The same survey explicitly calls out bankruptcy prediction as a promising area for causal analysis, suggesting that identifying the conditions under which a bank or firm goes bankrupt via causal methods could greatly enhance prescriptive analytics \cite{Kumar2023}. In short, there is both a need and an opportunity for innovative approaches that bring causal inference into corporate bankruptcy prediction to overcome the shortcomings of extant correlational models.

A persistent structural challenge in applying causal discovery to corporate bankruptcy data is the near-inevitability of violating causal sufficiency. Many economically critical determinants, such as management quality, strategic behavior, governance frictions, and market sentiment, remain unobserved or are only imperfectly proxied by available balance-sheet indicators. Classical causal-discovery methods typically assume that all relevant confounders are measured, an assumption rarely satisfied in corporate and financial datasets. As a consequence, data-driven graphs often become unstable, non-identifiable, or sensitive to small perturbations, especially when latent factors play a significant role in shaping firm performance and distress.
This limitation has motivated a growing interest in approaches that can explicitly acknowledge unobserved confounding, integrate proxy variables grounded in economic theory, and combine statistical testing with structured domain reasoning. In bankruptcy analysis, where key drivers of firm failure are partly unobserved and only partially captured by financial ratios, such hybrid strategies are essential for recovering robust, interpretable causal mechanisms suitable for counterfactual analysis.

For these reasons, this study introduces a novel approach to bankruptcy prediction that marries data-driven causal discovery with the reasoning capabilities of advanced AI. In particular, we propose ARCADIA (Agentic Reasoning for CAusal DIscovery Algorithm), an iterative causal DAG discovery framework powered by an agentic large language model (LLM). The LLM is employed as an autonomous ``research assistant'' that cycles through proposing, testing, and refining a causal graph in light of both domain knowledge and statistical evidence. This agentic workflow is a departure from conventional one-shot statistical algorithms; instead, the LLM operates in a loop of hypothesis generation and evaluation, allowing it to incorporate insights from financial theory, adapt to feedback, and progressively improve the causal model. The process is structured and statistically grounded: at each iteration, the LLM's suggestions (e.g. adding, removing, or orienting an edge in the DAG) are evaluated against data using rigorous tests or scoring metrics, ensuring that any refinements are supported by empirical evidence. By leveraging an LLM's ability to parse context (such as textual financial reports or prior research) and reason analogically, we systematically inject domain-informed priors into the causal discovery process. This approach represents a new direction at the intersection of causal inference and AI. It is not merely applying an out-of-the-box ML model to a finance problem, but rather using an AI agent to orchestrate the scientific modeling process itself. To our knowledge, this is one of the first attempts to harness AI reasoning and tool-use capabilities for causal modeling in economics. Through this fusion of an LLM with causal DAG discovery, our goal is to produce bankruptcy prediction models that are not only predictive, but also explanatory and robust – providing insight into the underlying causal structure of corporate distress. In the remainder of this paper, we first review the relevant literature and the theoretical underpinnings of our approach, then detail the methodology of the agent-based causal discovery process, and finally present empirical results on corporate bankruptcy data and discuss the implications for both causal inference research and financial risk management.

\section{Literature Review}

\subsection{LLMs for Causal Reasoning and Discovery}

LLMs have demonstrated impressive reasoning capabilities in recent years, prompting researchers to explore their utility beyond pure language tasks and into the realm of causal inference. Traditionally, causal discovery (CD) and LLMs have mainly evolved independently, but their convergence offers unique opportunities to advance causal understanding \cite{Wan2024}. Wan et al. (2024) survey this emerging synergy, identifying three main ways LLMs contribute to causal discovery: (1) direct inference - where LLMs infer causal graphs or subgraph structures directly from natural language descriptions and domain knowledge, (2) posterior correction - where LLMs validate and refine causal relationships identified by statistical causal discovery methods, and (3) prior knowledge - where LLMs serve as comprehensive sources of domain knowledge and contextual constraints for traditional causal discovery algorithms.
Early work along these lines treats the LLM as a proxy domain expert. For instance, researchers have used LLMs to answer conditional-independence queries during causal graph search, effectively replacing or supplementing human expertise in constraint-based algorithms. In the chatPC method \cite{Cohrs2024}, an LLM was prompted with questions of the form ``\textit{Are X and Y independent given Z?}'' for variables with known causal relations, and its answers drove the PC algorithm's edge removal decisions. The results showed promise – the LLM-aided PC procedure recovered a plausible causal graph on a complex real-world dataset – but also highlighted variability in the LLM's accuracy. Notably, the authors introduced a statistical aggregation method to combine multiple LLM query responses, thereby improving reliability by controlling for false positives and negatives. This exemplifies a broader point: while LLMs contain a wealth of latent knowledge (e.g., about typical economic cause-and-effect relations) and can perform chain-of-thought reasoning, they are not infallible and often benefit from structured approaches (like voting or consistency checks) when used for scientific inference. Recent studies have also begun to evaluate LLMs on causal reasoning benchmarks such as CLADDER and CORR2CAUSE, finding that certain prompting strategies can enhance an LLM's performance on causal tasks \cite{Wan2024}. Multi-agent approaches like MAC have also shown promise in improving causal discovery by leveraging debate and consensus mechanisms among multiple LLM agents \cite{Le2024}. Overall, the literature suggests that LLMs can contribute to causal discovery – by generating hypotheses, inferring qualitative causal directions, or serving as conditional independence oracles – but that they work best in tandem with formal methods that ensure statistical validity.

\subsection{Statistical Approaches to Causal DAG Discovery}

Causal DAG (Directed Acyclic Graph) discovery has been an active research area in statistics and machine learning for decades. The goal is to infer a DAG that represents the data-generating causal structure, using observational (and sometimes interventional) data. Glymour, Zhang, and Spirtes (2019) provide a comprehensive review of the field, covering the major families of approaches \cite{Glymour2019}. We briefly summarize these approaches and their evolution in recent years.

\subsubsection{Constraint-Based Algorithms} These methods rely on conditional independence tests to infer the graph structure. A classic example is the PC algorithm \cite{Spirtes2000}, which systematically removes edges that are ruled out by detected independencies. Under assumptions like the Causal Markov condition, faithfulness, and no latent confounders, PC is guaranteed to identify the true graph up to Markov equivalence \cite{Spirtes2000}. Variants such as FCI extend this approach to handle hidden confounders \cite{Spirtes2000}. Constraint-based methods are attractive for their interpretability and sound theoretical guarantees, but they can be brittle in practice: they require accurate independence tests (difficult with limited data or complex distributions) and can suffer from error propagation (early testing mistakes lead to cascading mis-specifications). As noted earlier, they also assume causal sufficiency, which may not hold in economic data \cite{Spirtes2000}.
    
    \subsubsection{Score-Based and Search Algorithms} These approaches formulate causal discovery as an optimization problem. They define a score (e.g., Bayesian score, BIC, or another goodness-of-fit metric penalized for complexity) for candidate DAGs and then search for the highest-scoring graph. Prominent algorithms include Greedy Equivalence Search (GES) and its variations, which add or remove edges in a way that improves the score until a local optimum is reached. Score-based methods can naturally incorporate priors and handle uncertainty by averaging over high-scoring graphs, at the cost of solving a generally NP-hard search problem. In recent years, there have been notable advancements in making this search more tractable. One breakthrough was formulating the structure search as a continuous optimization problem – for example, Zheng et al. (2018) introduced the NOTEARS algorithm, which represents the DAG adjacency matrix with differentiable constraints to enforce acyclicity \cite{Zheng2018}. By using gradient-based optimization, NOTEARS and its successors avoid exhaustive graph enumeration and can scale to larger networks. Such continuous relaxations have enabled the application of deep learning techniques (e.g. neural networks to parameterize causal links) within causal discovery \cite{Zheng2018}. Nonetheless, score-based methods require careful regularization to avoid overfitting, and the solution may only identify a Markov equivalence class of graphs rather than a unique DAG, especially in the absence of interventions.
    
    \subsubsection{Functional Causal Models and Hybrid Methods} A third category assumes specific functional forms for causal relationships and exploits structural asymmetries to orient edges. Examples include the LiNGAM algorithm, which assumes linear non-Gaussian dependencies, or nonlinear additive noise models. These methods often use independence of regression residuals or other patterns to infer causality (e.g., if $Y=f(X)+\text{noise}$ with independent noise, one can deduce $X \to Y$). They can succeed where purely combinatorial methods fail, but require correct model assumptions. Recent work also blends approaches – for instance, using an LLM or expert knowledge to constrain the search space of a score-based algorithm (a hybrid of knowledge-driven and data-driven search). The integration of domain knowledge is particularly relevant in economics, where we might rule out or favor certain causal links a priori. In fact, leveraging external knowledge to guide statistical discovery is a recurring theme in the literature \cite{Glymour2019}, and it is a principle that underpins our agentic LLM methodology.

Causal DAG discovery has evolved into a rich toolbox of algorithms. Each approach has strengths and limitations: constraint-based methods are principled but sensitive to assumptions, score-based methods are flexible but computationally intensive, and functional model approaches exploit structure at the expense of generality. The state-of-the-art increasingly involves combining these techniques or augmenting them with auxiliary information to improve reliability. Our work follows this trend by using an AI agent to inject domain-informed constraints and to adaptively steer the search, effectively serving as an intelligent heuristic on top of traditional statistical methods.

\subsection{Causal Modeling in Finance and Economics}

Causal inference is gaining traction in finance and economics as researchers seek models that not only predict outcomes but also inform policy and decision-making. Compared to fields like epidemiology or social sciences, however, the uptake of causal methods in finance has been relatively slow \cite{Kumar2023}. Conventional economic modeling has long included structural approaches (such as simultaneous equation models or vector autoregressions with imposed identification restrictions), but these typically rely on human-specified structures. Automated causal discovery in the financial domain is still novel. There are a few notable efforts pointing in this direction. For example, Cao et al. (2022) developed a two-stage Bayesian Network model for corporate bankruptcy prediction, where they first used LASSO to select key financial ratios and then learned a Bayesian network topology among those variables \cite{Cao2022}. This approach yielded performance on par with state-of-the-art machine learning models, while providing a transparent graphical representation of how various risk factors jointly influence default probabilities \cite{Cao2022}. The interpretability of such a causal model is a major advantage for investors and policymakers, as it lays out a chain of reasoning (e.g., ``leverage influences liquidity, which in turn affects bankruptcy risk'') rather than just an inscrutable prediction score. Another line of research focuses on macro-financial causality: for instance, Stolbov and Shchepeleva (2020) applied multivariate causal inference to assess how systemic risk and economic policy uncertainty impact firm bankruptcy rates \cite{Stolbov2020}. By using techniques akin to Granger causality and structural equation modeling, they provided evidence of directional influences (e.g., credit market shocks causing downstream increases in bankruptcies). These studies underscore the value of causal modeling in finance – offering scenario analysis, stress testing, and policy evaluation capabilities that purely predictive models lack.

Despite these initial forays, recent reviews characterize causal analytics in finance as an emerging frontier. Kumar et al. (2023) surveyed applications of causal inference in banking, finance, and insurance and concluded that the field is still in its early stages, with much room for innovation \cite{Kumar2023}. They emphasize that key financial problems (credit scoring, default prediction, fraud detection, etc.) could benefit from causal methods to enhance explainability and fairness. Indeed, regulators and industry practitioners have started to push for ``explainable AI'' in risk modeling, especially in high-stakes domains like lending where understanding why a model makes a prediction is crucial. Causal DAGs and counterfactual analysis present a natural path to such explainability. 

Our work positions itself in this nascent intersection of causal inference and financial modeling. By focusing on bankruptcy prediction – a problem with rich data and significant economic importance – we aim to demonstrate how an advanced causal approach can complement and out-explain traditional models. In doing so, we also respond to calls in the literature for new methodologies that make causal modeling more scalable and accessible to financial analysts \cite{Kumar2023}. If successful, the proposed agent-driven causal discovery framework could be extended to other economic forecasting tasks (e.g., market crash prediction, macroeconomic policy impact analysis), thereby broadening the toolkit for economists seeking data-driven yet theory-informed insights.

\subsubsection{Agentic AI and Iterative Causal Discovery}

Our proposed methodology draws inspiration from the concept of agentic AI: AI systems that proactively plan, execute, and refine actions towards a goal, often in a loop of continuous improvement. In the context of scientific research and discovery, there is a growing body of work on using AI agents to automate parts of the scientific method. For example, ResearchAgent is a system that leverages LLMs to assist researchers by generating novel research ideas, proposing methodologies, and even designing experiments in an iterative manner  \cite{ResearchAgent2024}. Starting from a seed topic, ResearchAgent retrieves relevant literature, formulates hypotheses, and then employs multiple LLM-based ``reviewer'' agents to critique and refine those hypotheses – mimicking the peer-review feedback loop to converge on robust research proposals \cite{ResearchAgent2024}. More recently, The AI Scientist-v2 extends this concept by eliminating the need for human-authored code templates and using agentic tree search to manage experiments across different machine learning domains. When tested, one of three AI-generated manuscripts submitted to an ICLR workshop exceeded the average human acceptance threshold for that venue \cite{Yamada2025}.  This showcases how LLMs, endowed with agent-like autonomy, can navigate complex problem spaces and improve solutions through iteration. Another compelling example is LLM-SR, a novel framework that leverages large language models for scientific equation discovery. This approach treats equations as executable programs and combines LLMs' scientific knowledge with evolutionary search algorithms. The system operates through an iterative refinement loop: the LLM generates equation program skeletons based on domain knowledge, these programs are then optimized against observed data using differentiable parameter optimization, and the best-performing equations are stored in a dynamic experience buffer that guides subsequent iterations. This continuous feedback cycle enables LLM-SR to efficiently navigate the vast equation search space, ultimately discovering physically accurate and interpretable mathematical models that significantly outperform traditional symbolic regression methods, particularly in out-of-domain generalization tasks \cite{shojaee2025llmsrscientificequationdiscovery}.

In the specific domain of causal discovery, the idea of agentic workflows is just beginning to take shape. Le et al. (2024) recently proposed a multi-agent framework for causal discovery in which distinct LLM agents take on specialized roles \cite{Le2024}. In their approach, Meta Agents engage in reasoning and debate to hypothesize causal structures, while Coding Agents generate and execute code to statistically evaluate those structures (using libraries for constraint-based or score-based learning). A hybrid model combines these, allowing the system to benefit from both the LLM's broad knowledge and the precision of numerical algorithms. The initial findings are encouraging: the multi-agent system effectively utilized LLMs' expert knowledge and reasoning to guide the search, and by cross-verification among agents, it mitigated some pitfalls of relying on a single model's judgment. This aligns with the broader theme in agentic AI that multiple perspectives (or agents) plus an ability to act in the world (e.g., run code, query data) can produce more reliable outcomes than a single-pass, static AI prediction. Our work can be seen as part of this emerging paradigm. We design an agentic process in which a single, versatile LLM (augmented with tool use for statistical computations) iteratively interacts with the data and a causal model. The LLM-agent generates hypotheses about how variables might be causally connected, then calls statistical routines (e.g. to compute likelihood scores or perform independence tests) to verify these hypotheses, and finally updates its beliefs or model description accordingly. This closed-loop setup continues until convergence. Such an approach draws on the strengths of both human-like reasoning and machine precision: the LLM contributes adaptability, creativity, and domain familiarity, while the statistical tools enforce discipline and objectivity in the causal search. We note that agentic AI in finance introduces additional considerations (e.g., ensuring the AI's actions remain within realistic and ethical bounds), but in a simulation or modeling context like ours, these concerns are manageable. The key takeaway from recent literature is that iterative, agent-driven discovery is a powerful strategy for complex problems – and we harness that insight here to tackle the complexity of causal bankruptcy modeling. By embedding an ``AI researcher'' in the modeling loop, we aim to push the frontier of both causal inference methodology and its application in economic analysis, illustrating a path forward for AI-assisted scientific discovery in theory-heavy domains.

\section{Methodology:  The ARCADIA Framework} \label{sec:agent-arch}

ARCADIA is an 
end-to-end, agentic workflow that discovers and validates causal 
relationships in tabular panel data.  
At its core, ARCADIA alternates between \emph{hypothesis generation}—posing a 
directed acyclic graph (DAG) that represents a causal theory—and 
\emph{statistical evaluation}, which tests that theory against data.  
The loop terminates once a DAG satisfies a predefined battery of causal  
diagnostics or once a budget of refinement rounds is exhausted.  All 
hyper-parameters are user-adjustable (Table~\ref{tab:params}), and the overall
control flow is summarised in Algorithm~\ref{alg:arcadia-main}.
ARCADIA is orchestrated as a four-node control graph that loops until a 
candidate DAG passes all causal diagnostics.  The workflow behaves 
differently in the \emph{first} iteration versus each \emph{refinement} 
iteration. The framework is built as a graph, with the following nodes.

\begin{table}[htpb]
 \centering
 \caption{Key hyper‑parameters of ARCADIA.  All values are user‑settable.}
 \label{tab:params}
 \begin{tabular}{@{}ll@{}}
   \toprule
   Symbol / Name & Meaning \\
   \midrule
   $k_{\text{init,min}}$, $k_{\text{init,max}}$   & Min./max.\ nodes in the initial DAG proposal \\

   $k_{\text{refine}}$                            & Max variables the agent may add/swap/remove per iteration \\
   $T_{\max}$                                     & Upper bound on refinement iterations \\
   $M$                                            & Number of columns sampled \\
   $\alpha$                                       & Nominal significance level for edge tests \\
   $\Theta_{\mathrm{global}}$                     & Threshold for the composite global‑validity score \\
   $\Theta_{R^{2}}$                               & Minimum average $R^{2}$ across node models \\
   $\Theta_{\mathrm{VIF}}$                        & Upper limit for variance‑inflation factors \\
   $T$                                            & Treatment variable(s) in the causal graph \\
   $Y$                                            & Outcome variable(s) of interest \\
   \bottomrule
 \end{tabular}
\end{table}

\paragraph{Node~1: \textsc{Initialise}.}
The full panel is read, and—if requested—a temporally balanced subset of columns is drawn to keep the search tractable. All hyper-parameters (Table~\ref{tab:params})  
        are recorded in a shared Graph State so subsequent nodes can reference a single 
        source of truth. A unique run identifier and an empty memory log 
        are created. In Algorithm~\ref{alg:arcadia-main}, this corresponds to
        the initial sampling of $V_M$, creation of the history log $\mathcal{H}$,
        and initialisation of the best-graph tracker.

\paragraph{Node~2: \textsc{Propose}.}
The behaviour of this node bifurcates:

\begin{description}
  \item[Iteration~1.]  The language model receives a 
        \emph{bootstrap prompt} that lists every available variable plus loose
        size bounds.  Lacking any historical feedback, the model must rely on
        domain priors to craft its initial causal hypothesis. The output is an initial DAG structured with three fields: \texttt{reasoning}, 
        \texttt{assumptions}, \texttt{edges} (see \S\ref{sec:refinement}). In Algorithm~\ref{alg:arcadia-main} this corresponds to the branch executed when $t=1$.
  \item[Iterations~$>\!1$.]  The prompt now includes the full diagnostic
        report of the most recent DAG, the rationale for its failure, and a
        strict budget of at most $k_{\text{refine}}$ variable changes.  
        Armed with this context, the model proposes a \emph{refined} DAG
        meant to fix specific shortcomings rather than start from scratch.
        This behaviour is captured by the refinement branch in Algorithm~\ref{alg:arcadia-main},
        where the failure memo and history $\mathcal{H}$ are fed back into the next proposal.
\end{description}

\paragraph{Node~3: \textsc{Evaluate}.}
Comprehensive statistical analyses are run on the DAG to provide the agent with data-driven feedback. If the DAG satisfies the six criteria listed in Table \ref{tab:evals}, the workflow jumps to \textsc{Finish}; otherwise, it loops back to \textsc{Propose} with an updated failure
        memo. The full evaluation logic, including structural checks, node- and
        edge-level statistics and global scores are implemented by the
        \textsc{EvaluateDAG} routine described in Algorithm~\ref{alg:arcadia-eval},
        which is invoked inside Algorithm~\ref{alg:arcadia-main}.

\paragraph{Node~4: \textsc{Finish}.}
All artefacts—DAG diagram, JSON transcript of every proposal and diagnostic, 
and a summary report—are written to persistent storage, after which the run 
terminates. This corresponds to the final block of Algorithm~\ref{alg:arcadia-main},
where the best graph $G^\star$ and its diagnostics are returned and logged.

\medskip
\noindent\textbf{Control flow summary.}\;
The resulting control graph is therefore 
\[
  \textsc{Initialise} \;\rightarrow\; 
  \bigl[\textsc{Propose}\;\rightarrow\;\textsc{Evaluate}\bigr]^{0{:}T_{\max}}
  \;\rightarrow\;\textsc{Finish},
\]
where the bracketed pair may repeat up to $T_{\max}$ times but can exit 
early if verification succeeds.  Iteration 1 uses a broad, data-driven prior; 
subsequent iterations exploit memory to execute focused, theory-driven 
repairs, steadily converging on a causally defensible graph.
This iterative
loop is made explicit in Algorithm~\ref{alg:arcadia-main}, while the statistics
used during \textsc{Evaluate} are summarised in Table~\ref{tab:stats} and
computed procedurally in Algorithm~\ref{alg:arcadia-eval}.

\begin{sidewaystable}[htpb]
  \centering
  \caption{Summary of the statistics computed during ARCADIA's \textsc{Evaluate} step.  
           Metrics are grouped by the graph component they diagnose.}
  \label{tab:stats}
  \begin{tabular}{@{}lll@{}}
    \toprule
    \textbf{Level}      & \textbf{Statistic}        & \textbf{Interpretation / Role in Diagnostics} \\
    \midrule
    Edge
      & Residual correlation $(\rho)$   & Strength of association after adjusting for other parents \\[2pt]
      & $p$‐value (raw \& FDR-adjusted) & Statistical support for the edge’s existence \\[2pt]
      & $\Delta\mathrm{BIC}$            & Evidence favouring the proposed direction over its reverse \\[2pt]
      & Coefficient estimate            & Magnitude and sign of the direct effect \\[2pt]

    Node
      & $R^{2}$ / pseudo‐$R^{2}$        & Goodness-of-fit of the node’s regression on its parents \\[2pt]
      & Adjusted $R^{2}$                & Fit penalised for model complexity \\[2pt]
      & Likelihood/F-test               & Joint significance of all parent coefficients \\[2pt]
      & Corresponding $p$‐value         & Statistical support for the node model \\[2pt]
      & VIF per parent                  & Multicollinearity diagnostic among parent variables \\[2pt]
      & Binary vs.\ continuous fit      & Chooses logit or OLS depending on the node’s scale \\[4pt]
    
    Global
      & Significant‐edge ratio          & Fraction of edges with adjusted $p < \alpha$ \\[2pt]
      & Significant-model ratio         & Fraction of nodes whose parent set is jointly significant \\[2pt]
      & Direction accuracy              & Share of edges with $\Delta\mathrm{BIC}>2$ \\[2pt]
      & Mean $R^{2}$                    & Average explanatory power across all node models \\[2pt]
      & Composite validity score        & Mean of the above three global indicators \\ 
    \bottomrule
  \end{tabular}
\end{sidewaystable}

\begin{algorithm}[htbp]
\footnotesize
\caption{ARCADIA: Agentic Causal DAG Discovery (Main Loop)}
\label{alg:arcadia-main}
\DontPrintSemicolon

\KwIn{Panel data $D$, variables $V$, treatment $T$, outcome $Y$, budget $M$, hyperparameters $(k_{\text{init,min}}, k_{\text{init,max}}, k_{\text{refine}}, T_{\max}, \alpha, \Theta_{\mathrm{global}}, \Theta_{R^{2}}, \Theta_{\mathrm{VIF}})$}
\KwOut{Validated DAG $G^\star$ with diagnostics, or best candidate}

Sample temporally balanced subset $V_M \subseteq V$ with $|V_M| = M$ (including $T$, $Y$). \;
Initialise memory log $\mathcal{H} \gets \emptyset$, iteration $t \gets 1$, best score $s^\star \gets -\infty$, best graph $G^\star \gets \emptyset$. \;

\While{$t \leq T_{\max}$}{
  \tcp{Node 2: Propose}
  \eIf{$t = 1$}{
    Query LLM with bootstrap prompt (columns $V_M$, size bounds, description of $D$, $T$, $Y$, temporal structure, assumption checklist) and obtain JSON with \texttt{edges}. \;
  }{
    Build failure memo from $\mathrm{Diag}_{t-1}$ and $\mathcal{H}$; query LLM with refinement prompt (previous DAG, diagnostics, failed criteria, limit $k_{\text{refine}}$) and obtain new \texttt{edges}. \;
  }
  Construct DAG $G_t$ from \texttt{edges} and append proposal plus prompt to $\mathcal{H}$. \;

  \tcp{Node 3: Evaluate}
  $(\mathrm{Diag}_t,\ \text{ok}) \gets \textsc{EvaluateDAG}(G_t, D, T, Y, \alpha, \Theta_{\mathrm{global}}, \Theta_{R^{2}}, \Theta_{\mathrm{VIF}})$. \;

  \tcp{Acceptance and tracking}
  \If{\text{ok}}{
    $G^\star \gets G_t$; $s^\star \gets \mathrm{Diag}_t.S_{\mathrm{global}}$; \textbf{break}. \;
  }
  \If{$\mathrm{Diag}_t.S_{\mathrm{global}} > s^\star$}{
    $G^\star \gets G_t$; $s^\star \gets \mathrm{Diag}_t.S_{\mathrm{global}}$. \;
  }

  Create a compact failure memo from $\mathrm{Diag}_t$ (which criteria failed and why) and store it in $\mathcal{H}$. \;
  $t \gets t + 1$. \;
}

\tcp{Node 4: Finish}
Persist $G^\star$, full transcript $\mathcal{H}$ and diagnostics for the run; return $G^\star$ and its diagnostics. \;

\end{algorithm}

\begin{algorithm}[htbp]
\footnotesize
\caption{\textsc{EvaluateDAG}: Structural and Causal Diagnostics}
\label{alg:arcadia-eval}
\DontPrintSemicolon

\KwIn{DAG $G$, dataset $D$, treatment $T$, outcome $Y$, thresholds $(\alpha, \Theta_{\mathrm{global}}, \Theta_{R^{2}}, \Theta_{\mathrm{VIF}})$}
\KwOut{Diagnostics $\mathrm{Diag}$, validity flag \text{ok}}

\tcp{Structural checks}
Enforce acyclicity on $G$; prune edges that violate temporal order; remove nodes disconnected from both $T$ and $Y$. \;
\If{$T$ or $Y$ missing}{
  Set $\text{ok} \gets \texttt{false}$; record structural failure in $\mathrm{Diag}$; \Return{$(\mathrm{Diag}, \text{ok})$}. \;
}

\tcp{Node-level models}
For each node $X_j$ with parents $\mathrm{Pa}(X_j)$: select regression type (OLS or logit), fit $X_j \mid \mathrm{Pa}(X_j)$ on $D$, compute $R^{2}$, adjusted $R^{2}$, joint significance and VIFs; store in $\mathrm{Diag}$. \;

\tcp{Edge-level statistics}
For each edge $(X_i \rightarrow X_j)$: compute residual correlation, edge coefficient, raw and FDR-adjusted $p$-values, and $\Delta \mathrm{BIC}$ for $X_i \rightarrow X_j$ vs.\ $X_j \rightarrow X_i$; store in $\mathrm{Diag}$. \;

\tcp{Global metrics}
From node- and edge-level results, compute significant-edge ratio, significant-model ratio, direction accuracy, mean node $R^{2}$, and composite validity score $S_{\mathrm{global}}$; store in $\mathrm{Diag}$. \;

\tcp{Positivity / overlap check}
If the minimal adjustment set $\mathcal{A}_{T \rightarrow Y}$ is non-empty, fit a logistic regression of $T$ on $\mathcal{A}_{T \rightarrow Y}$, compute propensity scores $\hat{e}(X)$ and the share of units with $0.05 \leq \hat{e}(X) \leq 0.95$; set a positivity flag depending on whether this share exceeds 90\%. Store the overlap metric and flag in $\mathrm{Diag}$. Otherwise, set the positivity flag to \texttt{true}. \;

\tcp{Causal identifiability and treatment effect}
Compute minimal back-door adjustment set $\mathcal{A}_{T \rightarrow Y}$; set identifiability flag depending on whether $\mathcal{A}_{T \rightarrow Y}$ exists. Fit treatment–outcome model (with $\mathcal{A}_{T \rightarrow Y}$ if non-empty), obtain $p$-value and $\Delta \mathrm{BIC}$ for $T \rightarrow Y$ vs.\ $Y \rightarrow T$; store in $\mathrm{Diag}$. \;

\tcp{Final decision}
Check simultaneously: (i) the back-door criterion holds and a valid (possibly empty) adjustment set $\mathcal{A}_{T \rightarrow Y}$ is identified; (ii) $\Delta \mathrm{BIC}_{T \rightarrow Y} > 0$; (iii) the treatment effect is statistically significant or explicitly justified as negligible; (iv) $S_{\mathrm{global}} \geq \Theta_{\mathrm{global}}$; (v) mean node $R^{2} \geq \Theta_{R^{2}}$; (vi) all VIF $\leq \Theta_{\mathrm{VIF}}$. (vii) the positivity flag is true (sufficient propensity-score overlap). \;
Set $\text{ok} \gets \texttt{true}$ if all pass, otherwise $\text{ok} \gets \texttt{false}$. \;

\end{algorithm}

\subsection{Causal Assumptions}\label{sec:assumptions}
Every proposed DAG must explicitly address the four core assumptions required for causal identification. First, regarding unobserved confounding, the agent must state which latent factors remain and how observed variables serve as proxies for these unobserved influences. Second, the positivity assumption requires that the agent argue no sub-population is locked into or out of the treatment, often demonstrated by reporting overlap in the propensity-score distribution across treatment groups. Third, the consistency assumption, also known as SUTVA, demands that the treatment definition must be invariant across units and free from interference effects between different units. Fourth, temporal ordering necessitates that parent variables must strictly precede their children in time, with year-over-year deltas anchored to the later year for proper causal sequencing.

If any assumption appears untenable during the analysis, the agent is instructed to flag the issue and either revise the DAG structure to address the violation or, if revision proves impossible, record the limitation in the run log for transparency and future reference.
If any criterion fails, the associated diagnostics feed back into the next 
\emph{Propose} step, guiding theory‑driven adjustments rather than blind 
metric optimisation.

\begin{sidewaystable}[htpb]
  \centering
  \caption{DAG Verification Criteria}
  \label{tab:evals}
  \begin{tabular}{@{}ll@{}}
    \toprule
    \textbf{Test Name} & \textbf{Requirement / Purpose} \\
    \midrule
    Graph validity & Both treatment and outcome variables must be present in the DAG \\[2pt]
                   & Ensures the causal question can be addressed \\[4pt]
    
    Identifiability & Back-door criterion yields valid minimal adjustment set for treatment-outcome pair \\[2pt]
                    & Ensures causal effect can be identified \\[4pt]
    
    Orientation & Directed edge from treatment to outcome favored over reverse (positive $\Delta\mathrm{BIC}$) \\[2pt]
                & Confirms correct causal direction \\[4pt]
    
    Edge significance & Treatment-outcome edge has $p < \alpha$ after multiplicity correction, \\[2pt]
                      & or negligible-effect explanation provided \\[2pt]
                      & Validates statistical significance of causal relationship \\[4pt]
    
    Global validity & Composite score averaging: (i) share of statistically supported edges, \\[2pt]
                    & (ii) share of significant node models, (iii) direction-of-causation accuracy; \\[2pt]
                    & must exceed $\Theta_{\mathrm{global}}$ \\[2pt]
                    & Ensures overall model quality and accuracy \\[4pt]
    
    Model adequacy & Mean $R^{2}$ across all node regressions $> \Theta_{R^{2}}$, and all VIF $< \Theta_{\mathrm{VIF}}$ \\[2pt]
                   & Guards against poor fit and multicollinearity \\[4pt]
    
                    & Ensures substantive breadth of the model \\
    \bottomrule
  \end{tabular}
\end{sidewaystable}

\subsection{Iterative Refinement Loop}\label{sec:refinement}

ARCADIA's iterative refinement phase implements a theory-driven causal discovery process that prioritizes scientific validity over statistical optimization. After evaluating the initial DAG, the system enters a propose-evaluate cycle where each iteration tests a coherent theoretical framework about the world's causal mechanisms. The agent receives accumulating contextual information including complete diagnostic summaries of all previous DAGs, user-supplied critiques based on failed criteria, and metadata constraining variable changes to at most $k_{\text{refine}}$ modifications per round. This conversational memory enables systematic exploration of competing causal theories rather than metric-driven search.

The refinement process follows a strict hierarchical decision framework that embodies the principle that discovering true causal structure supersedes engineering favorable statistics. At the highest priority, the agent addresses causal identification: when the minimal adjustment set is null, it tests alternative theoretical frameworks by adding major confounders or swapping variables to represent different causal mechanisms, treating non-identification as evidence that the current theory may be wrong rather than a problem to fix. Second, it examines theoretical validity by questioning whether orientation conflicts indicate reversed causation, bidirectional effects, or unmeasured common causes. Only after achieving identification does it consider effect detection, recognizing that insignificant treatment effects may reflect genuinely null relationships, measurement issues, or alternative causal pathways. Model diagnostics like multicollinearity or low $R^2$ receive lowest priority and are addressed only when improvements preserve theoretical coherence.

Variable modifications follow explicit theory-driven rules: additions test new causal mechanisms or proxy unobserved confounders, swaps evaluate competing theoretical frameworks, and problematic variables are retained if they appear in the minimal adjustment set or have strong theoretical justification. 
Each proposal is emitted as a JSON object with three mandatory fields:

\begin{description}
  \item[\texttt{reasoning}]  A narrative that justifies the chosen causal
        structure, cites financial theory, and lists the exact variable
        changes from the previous round.  This text serves as audible
        ``thinking'' for human reviewers.
  \item[\texttt{assumptions}]  A checklist‑based statement covering the four
        fundamental identification assumptions 
        (\S\ref{sec:assumptions}).  Explicit answers force the agent to
        confront potential violations early.
  \item[\texttt{edges}]  An array of ordered pairs that encodes the proposed
        DAG.  Because the evaluation engine is model‑agnostic, only the raw
        edge list is required.
\end{description}

The combination of structured output and rich free‑form explanation allows 
both automated scoring and qualitative inspection, closing the loop between
algorithmic discovery and substantive domain knowledge.

A distinctive methodological aspect of ARCADIA is that it does not rely on the assumption of causal sufficiency. Financial and corporate datasets inherently omit crucial factors such as managerial competence, governance quality, expectations formation, or strategic anticipation, meaning that purely data-driven causal discovery must operate under systematic latent confounding.
ARCADIA mitigates this challenge through three complementary mechanisms: (i) LLM-based reasoning proposes theory-grounded proxy confounders and alternative causal pathways; (ii) explicit verification of the back-door criterion ensures that the treatment–outcome effect is declared identifiable only when supported by a valid minimal adjustment set; and (iii) the iterative refinement loop tests multiple causal theories and corrects structural omissions by adding, removing, or swapping variables to restore identifiability.

\subsection{DAG Evaluation Methodology}

The ARCADIA framework employs a hierarchical evaluation system that operates at edge, node, and global levels, with statistics detailed in Table~\ref{tab:stats}. This multi-layered approach ensures that proposed DAGs meet both structural coherence and statistical validity requirements before proceeding through the iterative refinement process. 
Before statistical evaluation, ARCADIA performs essential structural checks that determine whether a DAG can proceed to causal analysis. The framework validates temporal ordering by removing edges that violate causal precedence, based on year extraction from variable names. Additionally, disconnected nodes—those with no causal path to the outcome—are identified and removed since they cannot contribute to confounding control or outcome prediction. A DAG is considered structurally valid only if both treatment and outcome variables remain present after these preprocessing steps. 
The evaluation proceeds through three integrated levels. Edge-level diagnostics assess individual causal relationships using partial correlations, FDR-corrected significance tests, and directional support via BIC comparisons. Node-level assessment evaluates each variable's regression model against its designated parents, automatically selecting appropriate model types and computing fit measures alongside multicollinearity diagnostics. These lower-level results synthesize into global validity measures that balance statistical significance, model adequacy, and directional accuracy, as summarized in Table~\ref{tab:stats}.
Beyond statistical assessment, the framework verifies causal identifiability using Pearl's back-door criterion and evaluates positivity through propensity score overlap analysis. These comprehensive diagnostics serve a dual purpose: establishing whether a DAG meets the verification criteria for acceptance, and providing targeted feedback to guide the LLM agent's structural modifications in subsequent iterations. Failed criteria generate specific error messages that inform theory-driven refinements, enabling the agent to systematically address validity concerns while maintaining causal coherence.

\section{Dataset}
The dataset utilized in this research was sourced from the AIDA database, which is maintained and distributed by Bureau van Dijk. This dataset comprises comprehensive financial data, including balance sheets, corporate demographic details, and product-related information, specifically for Italian non-financial firms. It explicitly excludes banks, insurance entities, and public sector bodies. The dataset includes Italian companies classified as medium or large according to criteria set forth by Legislative Decree 139/2015. Firms were selected if they surpassed at least two of three specified thresholds: net equity of 4 million EUR, net sales revenues of 8 million EUR, and an average of 50 employees during the fiscal year. Applying these criteria yielded a final dataset of 37,369 medium- to large-sized companies. This selection approach was specifically designed to yield a representative sample of businesses with substantial influence on Italy's broader economic landscape. Smaller firms were intentionally excluded due to their distinct economic dynamics and localized business challenges, which differ significantly from larger enterprises and could distort the analysis of macroeconomic or systemic financial phenomena. This rationale is supported by existing literature emphasizing medium and large enterprises as more reliable indicators of national economic conditions \citep{bancaditalia_bos_2025} 

Financial data covering the period from 2015 to 2019 provided an extensive array of economic indicators. Key financial measures include annual sales revenue, net profit, EBITDA, and total assets, all quantified in millions of euros. Profitability and operational efficiency were evaluated through ratios such as EBITDA-to-sales, Return on Equity (ROE), Return on Assets (ROA), Return on Investment (ROI), and Return on Sales (ROS). Additional financial structure indicators incorporated included net equity, short- and long-term debt ratios, fixed asset coverage ratios, current and liquidity ratios, and net financial position. Debt-related metrics encompassed financial charges relative to sales, interest coverage ratios, debt-to-EBITDA, debt-to-equity ratios, and the cost of borrowed funds. Furthermore, tax liabilities, both short- and long-term, were considered alongside the number of employees and company age, thereby providing additional insight into the companies' operational scale and maturity.

Temporal dynamics were captured by calculating year-to-year changes and differences over a three-year window for each financial indicator. Geographic information included province-level categorizations for major urban areas such as Milano, Roma, Brescia, and Torino, with low-frequency locations aggregated into a single category labeled ``other.'' Sectoral classifications employed the standard NACE codes, covering major industrial sectors and grouping less common sectors collectively. These comprehensive financial and demographic data yielded a final set of 150 features that capture both the static and dynamic aspects of corporate performance and financial health.

The target outcome variable was designed to indicate financial distress, coded \textit{1} if a company initiated bankruptcy proceedings in 2018 or 2019, and \textit{0} if it did not. Initially, the dataset exhibited a pronounced class imbalance, reflecting real-world conditions in which bankruptcy occurrences are rare compared to those of financially stable firms. To rectify this imbalance and ensure unbiased representativeness across industry sectors, a stratified undersampling technique was applied. The undersampling process matched the majority class observations to those of the minority class, maintaining proportional representation across different industrial sectors as identified by NACE codes. Consequently, the processed dataset comprises a balanced distribution with 434 total observations, equally split between financially distressed and healthy companies. Categorical variables, particularly those related to geography and sector classification, were encoded using one-hot encoding techniques to facilitate subsequent analytical processes. All financial figures within the dataset were uniformly expressed in millions of EUR, standardizing measurement scales and enhancing comparability across different companies and time periods.

\section{Experimental Setup}

All experiments were conducted on a MacBook Pro M3 Max equipped with 16 CPU cores and 64 GB of unified memory. The experimental framework was implemented in Python with dependencies detailed in the appendix, and a complete public GitHub implementation is provided for reproducibility. MLflow was utilized for experiment tracking and artifact management, with all runs logged to a local server for analysis.

We designed a comprehensive comparative evaluation to assess the performance of ARCADIA against established causal discovery baselines across varying problem complexities. The experimental design follows a factorial structure with two primary factors: algorithm type and variable subset size.

Four algorithms were evaluated in this study:
\begin{itemize}
\item \textbf{NOTEARS} - A continuous optimization approach that formulates causal discovery as a constrained optimization problem, using gradient-based methods to search for DAGs while enforcing acyclicity through a differentiable constraint.
\item \textbf{GOLEM} - A likelihood-based method that combines generalized linear models with continuous optimization, designed to handle both linear and nonlinear causal relationships while maintaining computational tractability.
\item \textbf{DirectLiNGAM} - A direct approach based on Linear Non-Gaussian Acyclic Models that exploits the statistical properties of non-Gaussian noise to identify causal directions without requiring iterative optimization.
\item \textbf{ARCADIA} - Our proposed agentic framework that combines large language model reasoning with statistical validation in an iterative refinement process to discover causally valid DAG structures.
\end{itemize}

All baseline algorithms were implemented using their default parameters as provided in the gCastle library to ensure fair comparison and reproducibility. For ARCADIA, we employed the GPT-5 model with standard hyperparameters: maximum 10 iterations, up to 5 variable changes per refinement, global validity threshold of 0.60, mean node $R^2$ threshold of 0.05, and VIF threshold of 10.0.

To evaluate algorithmic performance across different problem scales, we systematically varied the number of variables included in each causal discovery task. Three complexity levels were tested: $M \in \{20, 50, \text{150}\}$, where $M$ represents the total number of variables available to each algorithm for DAG construction, with $\text{150}$ indicating the use of the full dataset containing all 150 features. For subset sizes $M \in \{20, 50\}$, we employed a temporally balanced sampling strategy that categorizes variables into five temporal buckets: year-over-year changes from 2015 to 2016, year-over-year changes from 2016 to 2017, and static variables from years 2015, 2016, and 2017. For a target subset size $M$, the treatment and outcome variables are always included, while the remaining $M-2$ variables are distributed across temporal buckets with each bucket receiving $\lfloor(M-2)/5\rfloor$ variables minimum and the 2017 bucket receiving any remainder to ensure recent data representation. For example, with $M=20$, the allocation yields 2 mandatory variables (treatment and outcome), 3 variables each from the two delta buckets and years 2015 and 2016, and 6 variables from 2017. This design prevents temporal bias that could lead to spurious correlations rather than genuine causal relationships.

For subset sizes $M \in \{20,50\}$, we conducted 20 independent runs per algorithm by drawing different temporally balanced variable subsets; the same 20 subsets were used for all four algorithms to ensure comparability. In contrast, for the full feature set ($M = 150$) there is only a single possible subset—the complete set of 150 variables—so no further resampling over columns is possible. To still characterise the stochastic behaviour of ARCADIA’s agentic loop, we repeated the ARCADIA experiment 19 times on the full feature set using different random seeds (affecting, for example, LLM responses and internal sampling), whereas the baseline algorithms, which are deterministic conditional on the data and hyperparameters, were each run once. In total, this yields $4$ algorithms $\times$ $2$ subset sizes $\times$ $20$ replications, plus $19$ ARCADIA runs and one run per baseline at $M = 150$.
 
Each generated DAG was evaluated against a comprehensive battery of causal validity criteria designed to assess both statistical rigor and theoretical coherence. Structural validity verification ensured that both treatment (EBITDA margin delta 2015-2016) and outcome (bankruptcy 2018-2019) variables are present in the discovered DAG. Causal identifiability was assessed using Pearl's back-door criterion to determine whether the causal effect is identifiable given the proposed DAG structure, and minimal adjustment sets were computed. Directional support was evaluated using BIC comparisons ($\Delta$BIC) to assess whether the proposed treatment $\rightarrow$ outcome direction is statistically favored over the reverse. Statistical significance testing of edges used appropriate regression models (OLS for continuous, logistic for binary outcomes) with FDR correction for multiple comparisons. The global model quality employed a composite score based on the fraction of statistically significant edges, the proportion of adequate node models, directional accuracy, and overall explanatory power (mean node $R^2$). Multicollinearity assessment used Variance Inflation Factor (VIF) analysis to detect problematic collinearity among parent variables in each node's regression model.

Beyond statistical metrics, each DAG underwent validation for temporal consistency and causal assumptions. Temporal ordering was verified by ensuring that parent variables temporally precede their children, with special handling for year-over-year delta variables. Additionally, positivity was assessed through propensity score overlap analysis when adjustment sets were non-empty.

For each algorithm-subset size combination, we computed summary statistics across the 20 replications, including success rates (percentage of valid DAGs), average DAG characteristics (size, connectivity), and distributional summaries of quality metrics. This approach enables assessment of both central tendency and algorithmic reliability across different experimental conditions. The experimental framework was designed to provide a rigorous, reproducible evaluation of causal discovery performance while accounting for the inherent complexity and uncertainty in real-world economic data. All experimental runs were logged with complete provenance tracking, enabling detailed post-hoc analysis and verification of results.

\section{Results}\label{sec:results}

ARCADIA attains near-perfect valid-DAG rates in all settings—19/19 for $M=\text{150}$ and 20/20 for both $M=20$ and $M=50$—while requiring no temporal pruning at any budget (all zeros) and removing essentially no disconnected nodes (at most $0.350\pm0.686$ at $M=50$). By contrast, the baselines exhibit appreciable pruning. With the full feature set, NOTEARS shows large temporal violations (579 edges pruned) and many disconnected nodes (113), GOLEM shows 68 and 109 respectively, and DirectLiNGAM shows 134 and 121. On sampled subsets, pruning remains non-trivial for the baselines and generally grows with $M$: at $M=50$, temporal pruning averages $131.200\pm13.003$ for NOTEARS, $30.800\pm4.773$ for DirectLiNGAM, and $8.550\pm2.092$ for GOLEM; disconnected nodes pruned average $47.650\pm0.257$, $28.300\pm7.400$, and $13.550\pm1.360$, respectively. Valid-DAG rates reflect these patterns: DirectLiNGAM drops from 19/20 at $M=20$ to 15/20 at $M=50$, whereas NOTEARS reaches 20/20 at $M=50$ (19/20 at $M=20$) and GOLEM remains 20/20 at both $M=20$ and $M=50$.

ARCADIA consistently proposes mid-sized graphs—about 10 nodes and 20 edges on average: $(8.684\pm0.875,\ 18.105\pm3.945)$ at $M=\text{150}$, $(10.950\pm1.488,\ 22.300\pm3.977)$ at $M=20$, and $(9.750\pm1.688,\ 19.800\pm5.210)$ at $M=50$. DirectLiNGAM’s size varies with $M$ (e.g., $4.300\pm1.055$ nodes at $M=20$ vs.\ $15.900\pm6.760$ at $M=50$), while NOTEARS tends to produce relatively sparse graphs at $M=20$ and $M=50$ (nodes $\approx$ edges, e.g., $7.5$ vs.\ $7.6$ at $M=20$). GOLEM collapses to almost trivial structures on subsets ($\approx$2 nodes and $\approx$1 edge), consistent with its small edge counts.

Adjusted $R^{2}$ averages for ARCADIA are stable at roughly $0.10$ across budgets: $0.113\pm0.020$ ($M=\text{150}$), $0.097\pm0.018$ ($M=20$), and $0.105\pm0.023$ ($M=50$). DirectLiNGAM attains the highest fit on the full set ($0.232$; single run) and remains comparatively strong at $M=50$ ($0.125\pm0.046$), while NOTEARS and GOLEM yield substantially lower averages across subsets (NOTEARS: $0.021\pm0.011$ at $M=20$, $0.018\pm0.006$ at $M=50$; GOLEM: $0.014\pm0.000$ at $M=20$, $0.024\pm0.018$ at $M=50$).

End-to-end runtime (including ARCADIA’s proposal/evaluation loop) is on the order of several hundred seconds for ARCADIA: $496.818\pm157.047$~s ($M=\text{150}$), $879.388\pm198.510$~s ($M=20$), and $540.995\pm182.033$~s ($M=50$). Among baselines, DirectLiNGAM is fastest on subsets ($0.330\pm0.002$~s at $M=20$; $4.861\pm0.066$~s at $M=50$), whereas NOTEARS and GOLEM have higher and budget-dependent costs (e.g., NOTEARS $23.759\pm7.940$~s at $M=20$ and $193.256\pm11.327$~s at $M=50$; GOLEM $45.648\pm1.076$~s at $M=20$ and $104.108\pm2.057$~s at $M=50$).

\begin{figure}[htpb]
    \centering
    \includegraphics[width=0.7\linewidth]{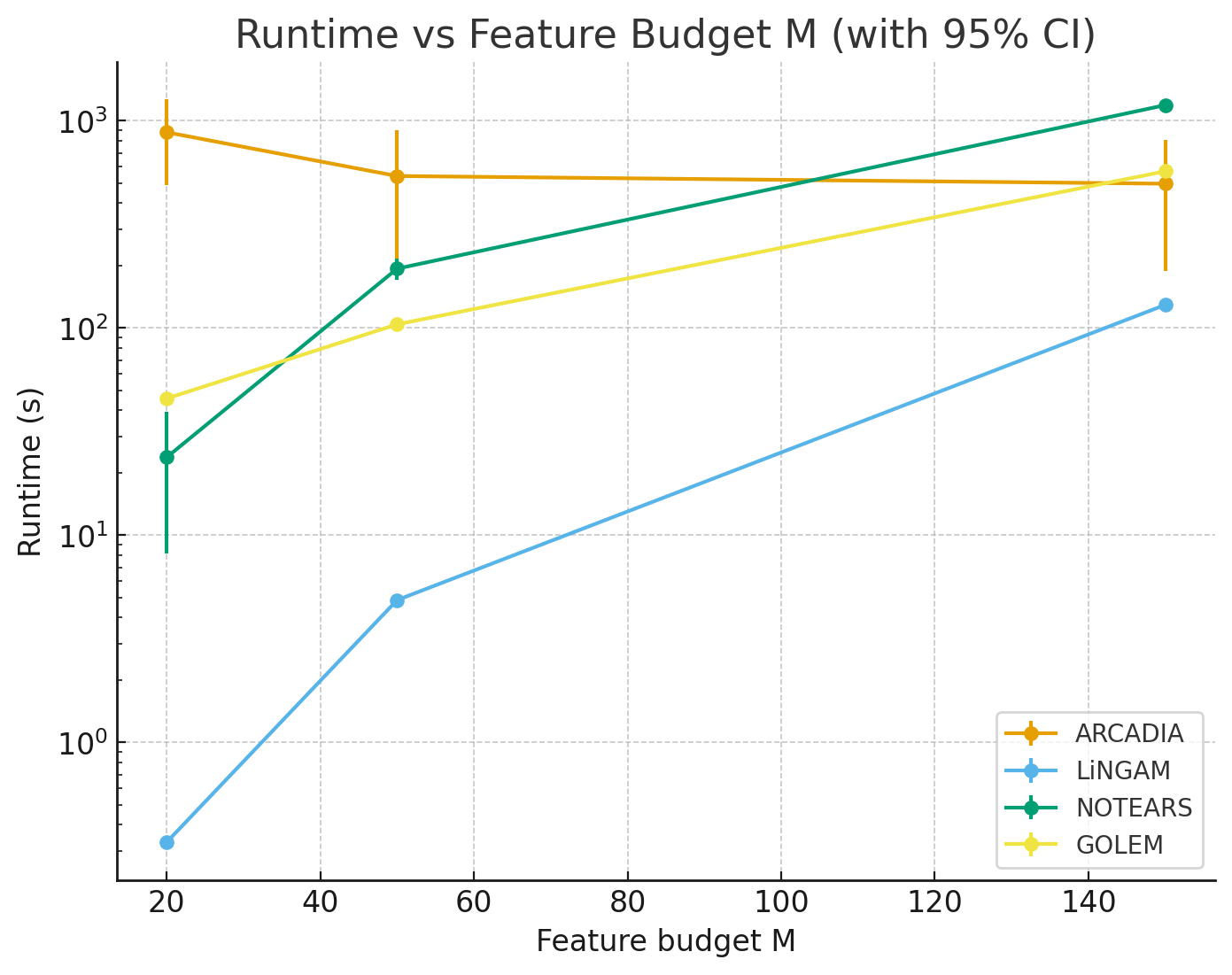}
    \caption{Runtime comparison across feature budgets $M$.}
    \label{fig:runtime}
\end{figure}

As shown in Figure~\ref{fig:runtime}, ARCADIA exhibits relatively higher runtimes at small feature budgets (\(M=20\)) compared to other methods. However, its scaling behavior with increasing \(M\) is more favorable. While competing approaches such as NOTEARS and GOLEM experience sharp increases in runtime as \(M\) grows, ARCADIA maintains a comparatively flatter growth curve, remaining efficient at larger feature budgets. DirectLiNGAM, in contrast, is very fast at small \(M\) but its runtime rises steeply when the number of features increases. These results indicate that ARCADIA achieves a balance between overhead at initialization and long-term computational efficiency.

Across budgets, ARCADIA (i) passes structural checks in essentially all runs, (ii) avoids temporal violations altogether, (iii) retains connected graphs of moderate size, and (iv) achieves average mean node adjusted $R^{2}$ values around $0.10$. Baselines vary by method and $M$: DirectLiNGAM offers comparatively higher fit at larger $M$ but with lower validity at $M=50$ and notable pruning; NOTEARS attains high validity by $M=50$ but with extensive pruning and low mean node adjusted $R^{2}$; GOLEM yields structurally valid but extremely small graphs on subsets with correspondingly low fit.

\begin{sidewaystable}[htpb]
\centering
\scriptsize
\caption{Performance vs.\ feature budget $M$ (mean $\pm$ sd across runs).}
\label{tab:all_M}
\begin{threeparttable}
\resizebox{\linewidth}{!}{%
\begin{tabular}{l l c c c c c c c}
\toprule
\textbf{Method} & \textbf{$M$} & \textbf{Number of nodes} & \textbf{Number of edges} & \textbf{Valid DAGs\tnote{a}} & \textbf{Temporal edges pruned\tnote{b}} & \textbf{Disconnected nodes pruned\tnote{c}} & \textbf{Runtime (s)\tnote{d}} & \textbf{Mean Node Adj. $R^2$\tnote{e}} \\
\midrule
ARCADIA & 20   & \(10.950 \pm 1.488\) & \(22.300 \pm 3.977\)  & 20/20 & \(0.000 \pm 0.000\)   & \(0.200 \pm 0.305\)   & \(879.388 \pm 198.510\) & \(0.097 \pm 0.018\) \\
ARCADIA & 50   & \(9.750 \pm 1.688\)  & \(19.800 \pm 5.210\)  & 20/20 & \(0.000 \pm 0.000\)   & \(0.350 \pm 0.686\)   & \(540.995 \pm 182.033\) & \(0.105 \pm 0.023\) \\
ARCADIA & 150 & \(8.684 \pm 0.875\)  & \(18.105 \pm 3.945\)  & 19/19 & \(0.000 \pm 0.000\)   & \(0.000 \pm 0.000\)   & \(496.818 \pm 157.047\) & \(0.113 \pm 0.020\) \\
\midrule
DirectLiNGAM  & 20   & \(4.300 \pm 1.055\)  & \(3.150 \pm 0.820\)   & 19/20 & \(6.300 \pm 1.372\)   & \(9.950 \pm 1.337\)   & \(0.330 \pm 0.002\)     & \(0.072 \pm 0.028\) \\
DirectLiNGAM  & 50   & \(15.900 \pm 6.760\) & \(22.050 \pm 12.334\) & 15/20 & \(30.800 \pm 4.773\)  & \(28.300 \pm 7.400\)  & \(4.861 \pm 0.066\)     & \(0.125 \pm 0.046\) \\
DirectLiNGAM  & 150 & \(6.000 \pm 0.000\)  & \(6.000 \pm 0.000\)   & 1/1   & \(134.000 \pm 0.000\) & \(121.000 \pm 0.000\) & \(129.172 \pm 0.000\)   & \(0.232 \pm 0.000\) \\
\midrule
NOTEARS & 20   & \(7.500 \pm 1.345\)  & \(7.600 \pm 3.402\)   & 19/20 & \(26.950 \pm 5.080\)  & \(16.550 \pm 1.840\)  & \(23.759 \pm 7.940\)    & \(0.021 \pm 0.011\) \\
NOTEARS & 50   & \(12.400 \pm 1.058\) & \(11.400 \pm 1.058\)  & 20/20 & \(131.200 \pm 13.003\)& \(47.650 \pm 0.257\)  & \(193.256 \pm 11.327\)  & \(0.018 \pm 0.006\) \\
NOTEARS & 150 & \(26.000 \pm 0.000\) & \(25.000 \pm 0.000\)  & 1/1   & \(579.000 \pm 0.000\) & \(113.000 \pm 0.000\) & \(1188.926 \pm 0.000\)  & \(0.016 \pm 0.000\) \\
\midrule
GOLEM   & 20   & \(2.000 \pm 0.000\)  & \(1.000 \pm 0.000\)   & 20/20 & \(2.200 \pm 0.661\)   & \(4.900 \pm 0.803\)   & \(45.648 \pm 1.076\)    & \(0.014 \pm 0.000\) \\
GOLEM   & 50   & \(2.050 \pm 0.098\)  & \(1.050 \pm 0.098\)   & 20/20 & \(8.550 \pm 2.092\)   & \(13.550 \pm 1.360\)  & \(104.108 \pm 2.057\)   & \(0.024 \pm 0.018\) \\
GOLEM   & 150 & \(6.000 \pm 0.000\)  & \(5.000 \pm 0.000\)   & 1/1   & \(68.000 \pm 0.000\)  & \(109.000 \pm 0.000\) & \(569.799 \pm 0.000\)   & \(0.090 \pm 0.000\) \\
\bottomrule
\end{tabular}%
}
\begin{tablenotes}\footnotesize
\item[a] Number of DAGs that contain both the designated treatment and outcome nodes (our validity criterion).
\item[b] Edges removed by temporal pruning (violations of time order).
\item[c] Nodes removed due to disconnection after pruning.
\item[d] End-to-end wall-clock time for the full pipeline per run.
\item[e] Mean across all node-level regressions in the DAG
\end{tablenotes}
\end{threeparttable}
\end{sidewaystable}

Table~\ref{tab:all_M} summarizes ARCADIA and three baselines under three feature-budget regimes: the full feature set ($M=\text{150}$) and two subset sizes ($M\in\{20,50\}$). The \emph{Valid DAGs} column reports, for each configuration, how many learned graphs include both the designated treatment and outcome nodes (our validity criterion) out of the total runs; \emph{Temporal edges pruned} and \emph{Disconnected nodes pruned} quantify post-hoc pruning required by the evaluator.

\section{Discussion}\label{sec:discussion}

Our empirical results (Section~\ref{sec:results}; Table~\ref{tab:all_M}) reveal a clear trade-off profile across methods. ARCADIA consistently returns structurally coherent graphs. Zero temporal violations and virtually no disconnected nodes—while maintaining mid-sized, interpretable DAGs with stable, moderate fit. In contrast, baseline learners frequently require heavy ex-post pruning to satisfy basic temporal and connectivity checks, and their graph size varies widely with $M$, occasionally collapsing to near-trivial structures (GOLEM) or oscillating between very small and very large graphs (DirectLiNGAM).

Three themes stand out. First, ARCADIA’s prompts (\S\ref{sec:agent-arch}; \S\ref{sec:refinement}) explicitly encode the 2015$\to$2017 temporal order and force the agent to reason about identifiability before chasing fit. As a result, the graphs it proposes already respect the temporal constraints the evaluation engine later enforces, so little to no pruning is needed. By contrast, the baselines discover structures first and are \textit{repaired} later by filtering. This difference matters: causal diagnostics (e.g., back-door sets, $\Delta\mathrm{BIC}$ orientation) are more trustworthy when applied to an ex-ante coherent graph rather than a heavily sanitized one. Second, ARCADIA repeatedly lands on DAGs with $\sim$10 nodes and $\sim$20 edges across $M\in{20,50}$ and the full feature set, suggesting that the agent’s theory-driven constraints regularize the search toward stable, mid-complexity models. Baselines either drift toward overly sparse (GOLEM on subsets) or fluctuate substantially with $M$ (DirectLiNGAM), with NOTEARS remaining valid but comparatively low-fit. In bankruptcy settings where interpretability and auditability are paramount, such stability is a feature rather than a bug. Last, ARCADIA's mean node-level adjusted $R^2$ hovers around 0.10—representing a deliberate balance where each node maintains meaningful connections to its parents while avoiding overfitting. This contrasts with approaches that maximize local fit (e.g., DirectLiNGAM at larger M) but may capture spurious associations. ARCADIA’s acceptance criteria (\S\ref{sec:assumptions}; Table~\ref{tab:evals}) prioritize identifiability, directionality, and diagnostic health over raw fit. The framework’s composite validity score, FDR-adjusted edge tests, $\Delta\mathrm{BIC}$ orientation checks, minimal adjustment sets are designed to prevent \textit{metric chasing} and to keep the agent focused on a causally defensible story rather than on maximizing explanatory variance.

Post-hoc pruning is not a benign clean-up. Trimming edges or nodes after seeing the data constitutes additional, unaccounted model selection; it can drop true confounders, change the estimand, bias effects, and undermine coverage unless the pruning step is explicitly modeled by a post-selection procedure \citep{Gradu2025JASA}. Our results therefore discourage ad-hoc post-pruning and instead advocate baking constraints into discovery, not after it.

Temporal ordering, presence of treatment/outcome, and identifiability checks are hard constraints, not afterthoughts. This yields graphs that are born valid rather than \textit{cleaned into validity}.
The refinement budget (at most $k_{\text{refine}}$ variable changes) forces the agent to articulate a coherent theoretical modification each round. This reduces the risk of data-driven tinkering that erodes interpretability. Each iteration logs a full rationale, a machine-checkable edge list, and diagnostic summaries. This is well-aligned with regulatory expectations for explainability in high-stakes finance.
The temporally balanced sampling and stable DAG sizes help ARCADIA remain effective as $M$ varies, preserving identifiability and temporal coherence even when the available feature set changes.

The runtime measurements reported here are based on remote API calls to a large language model. This overhead explains the higher absolute runtimes observed for ARCADIA at low feature budgets. Importantly, the scaling trend suggests that ARCADIA is better suited for scenarios with larger \(M\), where other algorithms become prohibitively expensive. In practice, runtimes could be substantially reduced by employing smaller models or optimized local inference, which would preserve ARCADIA’s advantageous scaling while alleviating its higher starting cost. Thus, although the absolute values may vary depending on implementation details, the relative efficiency patterns across methods are expected to remain consistent.

We emphasize that our reported $R^2$ values measure the average fit across all discovered causal relationships in the DAG, not the model's ability to predict bankruptcy. This design choice is intentional: optimizing for bankruptcy prediction accuracy would lead to correlation-focused models that may fail under intervention or changing economic conditions. By ensuring that each node in our causal graph is reasonably explained by its parents (even if only modestly), we maintain structural coherence while avoiding overfitting to spurious correlations. Future work could examine the trade-off between causal validity and predictive performance more systematically.

Our results highlight that ARCADIA remains robust even when key assumptions such as causal sufficiency are violated, a condition that characterizes virtually all corporate and financial datasets. Baseline causal-discovery methods tend to propagate structural errors when relevant confounders are unobserved, causing temporal violations, unstable orientations, and frequent pruning of disconnected components.
ARCADIA, by contrast, leverages LLM reasoning to introduce plausible proxy confounders, test competing theoretical mechanisms, enforce back-door identifiability, and evaluate global validity metrics. This enables the framework to compensate for latent confounding, similarly to an expert econometrician who integrates domain knowledge and diagnostic reasoning. As a result, ARCADIA yields coherent and interpretable causal graphs even when crucial economic drivers are imperfectly observed or entirely missing.

\paragraph{Limitations}

While ARCADIA demonstrates promising results for causal discovery in bankruptcy modeling, several limitations warrant discussion. First, the framework's reliance on large language models introduces methodological constraints. The quality of discovered DAGs depends on the LLM's embedded knowledge about financial relationships, which may reflect biases or outdated theories from its training data. The framework's performance is inherently bounded by the language model's understanding of economic causality, which cannot be directly validated or updated without retraining. Second, like any observational DAG-based approach, our framework relies on the assumption that relevant confounding can be sufficiently proxied by measured variables. Although ARCADIA explicitly encourages the agent to introduce theory-motivated proxies and checks identifiability via the back‑door criterion, unmeasured confounding remains a fundamental threat that cannot be entirely eliminated. This assumption is particularly strong in financial contexts where unobserved factors like management quality, market sentiment, strategic decisions, or informal business relationships play crucial roles. While ARCADIA attempts to address this through proxy variables and explicit assumption documentation, latent confounding remains a fundamental challenge that no data-driven approach can fully overcome. Third, ARCADIA's iterative refinement process introduces hyperparameter sensitivity. The choice of thresholds ($\Theta_{\mathrm{global}}$, $\Theta_{\mathrm{VIF}}$), refinement budget ($k_{\text{refine}}$), and maximum iterations ($T_{\max}$) can influence both the discovered structure and convergence behavior. Our experiments use fixed values across all runs, but optimal settings likely depend on the specific domain and dataset characteristics. The framework would benefit from principled hyperparameter selection methods or adaptive threshold adjustment based on data properties. Finally, our validation relies on a single bankruptcy dataset for Italian firms covering a specific period (2015-2019). The generalizability to other geographic regions, time periods, or financial distress definitions remains untested. Cross-market validation is critical given potential differences in bankruptcy laws, accounting standards, and business practices across jurisdictions. Moreover, the balanced sampling strategy (equal numbers of bankrupt/healthy firms) may not reflect the severe class imbalance encountered in practice, potentially affecting the causal structures discovered when applied to naturally imbalanced datasets.

\paragraph{Implications for bankruptcy modeling and financial risk.}
For practitioners, ARCADIA offers a concrete path from black‑box prediction to \emph{explainable, intervention‑ready} models. By outputting a vetted DAG along with minimal adjustment sets and sensitivity diagnostics, the framework supports counterfactual queries (“What if EBITDA margins improved by $\delta$?”), scenario analysis, and stress testing under policy‑relevant assumptions. Because the agent documents its theory at each step, risk teams gain a narrative linking firm fundamentals to distress—useful for governance, credit committees, and regulators. Methodologically, the study demonstrates that agentic LLMs can serve as orchestration layers that \emph{constrain} statistical discovery rather than loosen it, aligning flexible reasoning with strict identification criteria.

\section{Conclusion}\label{sec:conclusion}

We introduced ARCADIA, an agentic framework that couples LLM‑driven, theory‑aware hypothesis generation with a rigorous, multi‑layer evaluation of causal DAGs for bankruptcy analysis. Across variable budgets, ARCADIA produced temporally coherent, connected graphs with near‑perfect structural validity and stable, moderate fit—prioritizing identifiability, directionality, and diagnostic health over raw explanatory power. Baseline algorithms, while fast and sometimes higher‑fit in specific regimes, often required substantial ex‑post pruning or collapsed to overly sparse structures, underscoring the value of building \emph{causal guardrails} into the discovery loop itself. For the field, the key contribution is not a single accuracy figure but a \emph{process}: a reproducible, auditable way to discover causal structure that is constrained by economic theory and verified by data. This process aligns with emerging expectations for explainable AI in finance and provides a template adaptable to other domains where interventions—not just predictions—matter. The main costs are computational and design‑centric: agent loops introduce runtime overhead, and outcomes can be sensitive to prompting and thresholds. These are tractable engineering challenges. The larger research agenda is methodological: combining agentic reasoning with stronger orientation signals (e.g., invariance, quasi‑experiments), principled treatment of missingness, and cross‑environment validation. Taken together, our results suggest that agentic AI can help move corporate‑distress modeling from pattern recognition toward \emph{causal understanding}. If adopted widely, this shift would enable risk managers and policymakers to ask—and answer—the counterfactual questions that predictive models alone cannot address.

\section*{Declaration of Competing Interest}
The authors declare that they have no known competing financial interests or personal relationships that could have influenced the work reported in this paper.

\section*{CRediT Author Statement}
\textbf{Fabrizio Maturo}: Conceptualization, Methodology, Supervision, Writing - Review and Editing. 
\textbf{Donato Riccio}: Conceptualization, Software, Methodology, Formal Analysis, Writing - Review and Editing, Data Curation. 
\textbf{Andrea Mazzitelli}: Supervision, Validation, Writing - Original Draft. 
\textbf{Giuseppe Bifulco}: Data Curation, Resources, Writing - Review.  
\textbf{Francesco Paolone}: Data Curation, Resources, Writing - Review.  
\textbf{Iulia Brezeanu}: Conceptualization, Software, Methodology, Formal Analysis, Writing - Review and Editing, Data Curation.

\section*{Data Availability}
The AIDA dataset used in this study is proprietary and distributed under license by Bureau van Dijk.  
Due to licensing restrictions, the raw data cannot be publicly shared.

\section*{Funding}
This research was supported by Universitas Mercatorum under the FIN-RIC 2024-2025 research funding program for the project titled: 'AISECURE - AI Solutions for Enterprise Crisis Understanding and Risk Evaluation'.

\section*{Acknowledgements}
The authors thank Universitas Mercatorum for institutional support and the colleagues who provided valuable feedback during the development of this work.

\bibliographystyle{elsarticle-num}
\bibliography{sn-bibliography}

\appendix

\appendix
\section{Prompts}

\subsection{System prompt}

\begin{lstlisting}[breaklines=true, breakatwhitespace=true]

system_msg_initial = """
You are an autonomous causal-inference researcher.

**Objective** Estimate the causal effect of **{treatment} -> {outcome}** by proposing
an initial Directed Acyclic Graph (DAG).

Remember that {treatment} and {outcome} MUST BE IN THE DAG.

**Task**
Propose a starting DAG with between {initial_min_cols} and {initial_max_cols} nodes that represents
your best hypothesis about the causal relationships between the treatment, outcome,
and potential confounders or mediators. The graph must be acyclic.

---
### Causal-assumption checklist
In **your `assumptions` field** explicitly answer:
1. *Unobserved confounding*: which latent factors remain, and how do your chosen variables proxy for them?
2. *Positivity*: are there sub-groups where treatment is (almost) deterministic?
3. *Consistency / SUTVA*: why is {treatment} well-defined and why don't firms interfere?
4. *Temporal ordering*: confirm every parent variable predates its children (2015 -> 2017).
If any assumption looks doubtful, flag it.

"""

\end{lstlisting}

\subsubsection{Current user message}

\begin{lstlisting}[breaklines=true, breakatwhitespace=true]

current_user_msg = """
Iteration {iteration}: Propose a DAG with the following schema.

Return a JSON object exactly in the following schema.
Available columns: {all_cols_str}

MUST STRICTLY USE THE SAME COLUMN NAMES AS IN THE WORKING DATAFRAME.
YOU ARE ONLY ALLOWED TO ADD/REMOVE/SWAP COLUMNS A MAXIMUM OF {max_refinement_cols} TIMES.

Each record represents a medium-to-large Italian non-financial company with
2015-2017 financial-statement features.
Columns that start with "delta" are year-over-year changes.
`bankruptcy` = 1 if the firm filed for bankruptcy during 2018-2019, else 0.

Remember that {treatment} and {outcome} MUST BE IN THE DAG.

{
    "reasoning": "string",
    "assumptions": "string",
    "edges": [["parent", "child"], ["parent", "child"], ...],
}

"""

\end{lstlisting}

\subsubsection{Previous user message}

\begin{lstlisting}[breaklines=true, breakatwhitespace=true]

previous_user_msg = """
Iteration {iteration}: Propose a new DAG with the following schema.
YOU ARE ONLY ALLOWED TO ADD/REMOVE/SWAP COLUMNS A MAXIMUM OF {max_refinement_cols} TIMES.

Return a JSON object exactly in the following schema.

{
    "reasoning": "string",
    "assumptions": "string",
    "edges": [["parent", "child"], ["parent", "child"], ...],
}

"""

\end{lstlisting}

\subsubsection{Refinement}

\begin{lstlisting}[breaklines=true, breakatwhitespace=true]

system_msg_refinement = """

### Your task

Based on the historical feedback, propose a *refined* DAG that addresses the recurring issues.
* You may **add, swap, or remove up to {max_refinement_cols} columns** compared with the
  previous DAG in the past messages.
* Keep the graph acyclic and ensure the total node count stays within the
  allowed range.

Explain your causal reasoning **first**, output a new DAG with
a JSON object that matches the provided schema.

CAUSAL INFERENCE OBJECTIVE: Discover the true causal relationship between {treatment} and {outcome} by testing different DAG structures that reflect theoretical possibilities.

FUNDAMENTAL PRINCIPLE: You are DISCOVERING causal structure, not engineering it. Variable changes should reflect different theories about how the world actually works.

DECISION HIERARCHY:

PRIORITY 1: CAUSAL IDENTIFICATION
- If minimal_adj_set is null -> Test alternative theoretical frameworks:
  a) ADD a major confounder you may have missed (unobserved heterogeneity proxies)
  b) SWAP to variables representing different theoretical mechanisms
  c) If repeatedly null across reasonable theories -> Accept that identification may require external variation
- If minimal_adj_set exists -> These variables are REQUIRED (cannot be removed without theoretical justification)

PRIORITY 2: THEORETICAL VALIDITY
- If orientation_ok is false -> Consider alternative causal theories:
  a) Maybe the true causal direction is reversed
  b) Maybe there's bidirectional causation
  c) Maybe there's an unmeasured common cause
- DAG structure must reflect plausible real-world causal mechanisms

PRIORITY 3: EFFECT DETECTION AND PRECISION
- If treatment -> bankruptcy is insignificant -> Consider:
  a) True effect might be zero (not a problem to fix)
  b) ADD precision variables (reduce outcome noise)
  c) TEST different theoretical pathways (direct vs. mediated effects)
- Focus on identifying the effect that actually exists, not manufacturing significance

PRIORITY 4: MODEL DIAGNOSTICS
- High VIF: Accept if variables are causally necessary; avoid if redundant
- Low R^2: Accept if causal structure is sound; only improve if it aids identification

SPECIFIC DECISION RULES:

WHEN TO ADD A VARIABLE:
- Testing a new theoretical mechanism (e.g., "maybe leverage is the key confounder")
- Proxying for unobserved confounders (e.g., management quality via efficiency ratios)
- Improving precision of causal estimates (reducing noise)
- Exploring mediation pathways suggested by theory

WHEN TO SWAP A VARIABLE:
- Current variable contradicts theoretical expectations consistently
- Testing competing theoretical frameworks (e.g., size vs. efficiency as key factor)
- Current variable provides no causal information and theory suggests better proxy

WHEN TO RETAIN A PROBLEMATIC VARIABLE:
- It's in minimal_adj_set (causally required)
- Strong theoretical justification even if statistically weak
- Part of the causal mechanism you're investigating

VARIABLE SELECTION LOGIC:
1. What causal mechanism am I testing? (theory-driven)
2. What confounders does this theory require? (identification-driven)
3. How can I best measure these constructs? (measurement-driven)
4. NOT: How can I maximize statistical metrics?

HANDLING COMMON SCENARIOS:

NULL MINIMAL ADJUSTMENT SET:
-> "The current theory implies the effect isn't identifiable. Let me test a different theoretical framework."
NOT: "Let me add variables until it becomes identifiable."

INSIGNIFICANT TREATMENT EFFECT:
-> "Maybe the effect is truly small/zero, or I need better measurement."
NOT: "Let me add variables until it becomes significant."

FALSE ORIENTATION:
-> "Maybe my causal assumptions are wrong. Let me test the reverse direction."
NOT: "Let me add controls until orientation becomes true."

ITERATION FOCUS:
Each iteration should test a coherent theoretical story:
- "Theory A: Firm size is the key confounder"
- "Theory B: Operational efficiency drives everything"
- "Theory C: Industry dynamics dominate"

NOT: "Let me try adding this variable and see what happens"

**Verification criteria** All six conditions must hold simultaneously to declare the thesis VERIFIED:

1. *Identifiable* - `minimal_adjustment_set` is not null.
2. *Orientation* - treatment -> outcome has DeltaBIC > 0.
3. *Edge significance* - p-value < {alpha} (or a justified negligible effect).
4. *Global validity* >= {global_validity_threshold} and average R^2 >= {r2_threshold}.
5. *No multicollinearity* - every VIF <= {vif_threshold}.


### Causal-assumption checklist
In **your `assumptions` field** explicitly answer:
1. *Unobserved confounding*: which latent factors remain, and how do your chosen variables proxy for them?
2. *Positivity*: are there sub-groups where treatment is (almost) deterministic?
3. *Consistency / SUTVA*: why is {treatment} well-defined and why don't firms interfere?
4. *Temporal ordering*: confirm every parent variable predates its children (2015 -> 2017).

If any assumption looks doubtful, flag it.

REMEMBER: Your goal is scientific discovery of causal relationships, not optimization of statistical metrics. A well-identified small effect is more valuable than a poorly-identified large effect.

Reason on previous DAGs. What worked and what didn't? And propose a new DAG.

REMEMBER TO ADD/SWAP/REMOVE A MAXIMUM OF {max_refinement_cols} COLUMNS. You must list your changes as part of the reasoning, including a complete rationale based on financial theory AND statistical metrics.

Return a JSON object exactly in the following schema.

{
    "reasoning": "string",
    "assumptions": "string",
    "edges": [["parent", "child"], ["parent", "child"], ...],
}

"""

\end{lstlisting}

\appendix
\section{Implementation Details}

\subsection{Software Dependencies}
The experimental framework is implemented in Python 3.12+ using Poetry for dependency management. Key dependencies include MLflow (v3.1.0+) for experiment tracking, LiteLLM (v1.73.0+) for language model inference, gCastle (v1.0.4+) for baseline causal discovery algorithms, Statsmodels (v0.14.2+) for statistical evaluation, PGMPy (v1.0.0+) for probabilistic graphical models, and LangGraph (v0.4.8+) for agentic workflow orchestration. Additional dependencies comprise standard scientific computing libraries: Pandas (v2.3.0+), NumPy (v2.3.1+), SciPy (v1.11+), causal-learn (v0.1.4.3+), and LiNGAM (v1.10.0+) for specific causal discovery implementations.

\subsection{Data Availability and Preprocessing}
The AIDA database used in this study is distributed under a proprietary license by Bureau van Dijk and cannot be redistributed. However, the complete data preprocessing pipeline, including filtering criteria, feature engineering steps, missing value handling procedures, and stratified undersampling methodology, is provided in the public repository. Researchers with access to the AIDA database can reproduce the exact dataset used in our experiments by following the documented preprocessing steps.

\end{document}